# Development and Whole-Body Validation of Personalizable Female and Male Pedestrian SAFER Human Body Models

Natalia Lindgren[a]*, Qiantailang Yuan[a], Bengt Pipkorn[b], Svein Kleiven[a] and Xiaogai Li[a]

*[a]Division of Neuronic Engineering, KTH Royal Institute of Technology, Stockholm, Sweden; [b]Autoliv Research, Vårgårda, Sweden*

*Corresponding author (N. Lindgren)

*E-mail address*: nalindgr@kth.se

*Postal address:* KTH Flemingsberg, Hälsovägen 11C, 141 57 Huddinge

# Development and Whole-Body Validation of Personalizable Female and Male Pedestrian SAFER Human Body Models

## Abstract

Vulnerable road users are overrepresented in the worldwide number of road-traffic injury victims. Developing biofidelic male and female pedestrian HBMs representing a range of anthropometries is imperative to follow through with the efforts to increase road safety and propose intervention strategies. In this study, a 50$^{th}$ percentile male and female pedestrian of the SAFER HBM was developed via a newly developed image registration-based mesh morphing framework for subject personalization. The HBM and its accompanied personalization framework were evaluated by means of a set of cadaver experiments, where subjects were struck laterally by a generic sedan buck. In the simulated whole-body pedestrian collisions, the personalized HBMs demonstrate a good capability of reproducing the trajectories and head kinematics observed in lateral impacts. The presented pedestrian HBMs and personalization framework provide robust means to thoroughly and accurately reconstruct and evaluate pedestrian-to-vehicle collisions.

## Introduction

With non-proportionally high fatality and injury rates (WHO 2018), the safety of vulnerable road users remains a leading public health concern worldwide. Vulnerable road users, referring to e.g. pedestrians, cyclists and motorcyclists (OECD 1998), represent almost half of all traffic-related deaths (WHO 2018). With no indicators of progress in downward trends in the health burden (WHO 2018; NHTSA 2021; Ameratunga et al. 2006), there is an evident need of injury prevention measures.

Investigations of traffic accident situations have traditionally been carried out using full-scale anthropomorphic test devices (ATDs), i.e. crash test dummies. The widely used ATDs have been developed to mimic the anthropometry, articulation, and kinematic response of the human body in impacts, serving as a tool to predict potential human injuries in accidents (Crandall et al. 2011). However, ATD crash testing is a very time-consuming and expensive process, and full-scale pedestrian dummy tests are rarely imposed in current pedestrian safety test procedures. Most safety standards that encompass the pedestrian safety field have been concentrated on subsystem tests using impactors representing the most critical human body regions, such as the head or leg, even though it is debated to what extent these procedures represent real-world pedestrian accidents (Matsui et al. 2002; Ptak et al. 2012).

In the past decades, virtual human surrogates have been gaining more ground among researchers and in automotive industries. Computational reconstructions of destructive crash scenarios are often carried out using state-of-the-art Finite Element (FE) human body models (HBMs). Since the HBMs oftentimes are anatomically detailed, they can provide deep insight into the injury mechanisms of specific tissues during dynamic loading scenarios. As opposed to the physical dummies, HBMs also enable measurements of tissue-based metrics to be used for injury assessments (Crandall et al. 2011). A wide selection of HBMs are available for application, such as the widely used Total HUman Model for Safety (THUMS) (Iwamoto et al. 2002), the Global Human Body Models Consortium (GHBMC) (Decker et al. 2019) and the VIVA+ HBM (John et al. 2022). Another advocated HBM is the SAFER HBM (Pipkorn et al. 2021; Östh et al. 2021), that was originally based on the THUMS v3. The SAFER HBM represents a 50$^{th}$ percentile male occupant and is provided with lumped internal organs and an active muscle package that enables human postural control (Ólafsdóttir et al. 2019; Östh et al. 2015). The SAFER HBM includes an extensively validated head model that was originally developed and used for brain injury research (Kleiven 2002, 2007; Kleiven and Von Holst 2002). With the head being one of the most commonly injured body part among road users, as well as the major cause of death and hospitalization (OECD et al. 1998), this poses a great advantage of the SAFER HBM.

Traditionally, ATDs and HBMs have been developed to mirror the anatomy of an average, often times US American (Schneider et al. 1983), male. The recognition of the importance of including females and wider size ranges in traffic safety assessments is nevertheless increasing (Davis et al. 2016; Larsson et al. 2022). A far-reaching advantage with the HBMs is that their geometry can be modified to represent a broader anthropometric spectrum compared to the full-scale ATDs. Multifold methods for geometric subject-personalization have been used in association with the utilization of HBMs, with the most common approach being different forms of volumetric HBM scaling (Wu et al. 2017; Danelson and Stitzel 2015). The scaling processes can for instance introduce a length scaling factor to match the subject's stature and second scaling factor in the transverse plane to match the total bodymass. Another personalization option is parametric morphing of an HBM to a target geometry (Costa et al. 2020; Piqueras-Lorente et al. 2018; Larsson et al. 2022), which in many cases is one of the more cost and resource-expensive personalization option, especially since the proposed methods usually require the identification of landmarks and control points to obtain the later applied computational heavy mesh transformation. Image-registration-based morphing techniques that enable mesh morphing in a matter of minutes have been promoted in other fields of application (Li et al. 2021). This effective morphing strategy has also been shown to be effective for HBM personalization (Li et al. 2023). Correcting HBM geometries using morphing techniques has been shown to be the prevailing option for personalization in regard to maintaining biofidelity (Poulard et al. 2016).

Currently, the SAFER HBM is only available in a seated, occupant posture. Despite the fact that vulnerable road users are overrepresented in the worldwide number of road-traffic injury victims, most proposed intervention strategies have been centered on the protection of vehicle occupants, and little on the vulnerable road users (Ameratunga et al. 2006; Crandall et al. 2002). In that regard, vulnerable road users should be of substantial focus in current prevention strategies, and developing biofidelic pedestrian HBMs is imperative to follow through with these efforts. Including female versions, as well as supplying personalization pipelines along with the HBMs, are equally important since the anthropometry has shown to have a non-negligible influence in simulation accuracy (Larsson et al. 2019, 2022; Poulard et al. 2016). Evidence suggesting that females carry a higher risk of sustaining injuries during motor vehicle crashes compared to males (Ulfarsson and Mannering 2004; Evans and Gerrish 2001; Evans 2001), stresses this necessity even further.

The aim of this study is thus to develop a baseline male and female pedestrian version of the SAFER HBM, along with a robust framework for subject personalization. The study presents the application of a previously developed landmark-free image-registration-based morphing approach that allows rapid generation of subject-specific HBMs (Li et al. 2023). The HBM biofidelity will be evaluated in whole body impacts using a generic sedan buck. The HBMs are targeted for vehicle-to-pedestrian impact simulations, with an emphasis on head kinematics.

## Materials and Methods

All simulations referred to in this study were pre-processed using LS-PrePost v4.8 and PRIMER v18.1 (Oasys Ltd, London, UK). Simulations have been performed using LS-DYNA v13 (LSTC, Livermore, CA, US) with multiple CPUs, while post-processing was carried out using MATLAB v2021a (The MathWorks, Inc., Natick, MA, United States). An overview of the study is provided in Figure 1.

### *Development of Baseline Pedestrian HBMs*

To obtain two baseline pedestrian HBMs, the SAFER occupant HBM was morphed to have the shape and the inner skeleton geometry representative of an upright, standing 50$^{th}$ percentile male and female. This was done in four subsequent processing steps: (1) the preparation of target geometries, (2) the pre-processing of the occupant HBM to obtain a template geometry, (3) the image-registration based morphing procedure and (4) the personalization accuracy evaluation.

#### *Target geometries*

Target geometries were generated using three main tools: Skinned Multi-Person Linear Models

(SMPLs) (Loper et al. 2015), SMPL eXpressive (SMPLX) (Pavlakos et al. 2019), and the Obtaining Skeletal Shape from Outside (OSSO) system (Keller et al. 2022). SMPLs are skinned vertex-based models of the human body that represents a wide range of body shapes in natural human poses. The body models were originally generated based on input from the CEASAR dataset (Blackwell et al. 2002), which compiles body scans of over 4,000 human subjects. Unlike the SMPLs, SMPLX also enables the generation of per-joint pose corrective blendshapes that depend on both body pose and BMI. OSSO maps an internal skeleton to a 3D body surface. The OSSO generated skeletons are based on 2,000 processed medical images of human subjects from the UK Biobank (Sudlow et al. 2015).

An outer skin surface model was generated using SMPLX by using height, weight and gender as input. The 50$^{th}$ percentile male and female have been reported to have a height and weight of 175 cm and 77 kg, and 162 cm and 62 kg, respectively (Schneider et al. 1983). The SMPLX models were positioned to an upright standing position using Blender v3.4 (The Blender Foundation, Amsterdam, Netherlands) and an available plugin (www.github.com/Meshcapade/SMPL_blender_addon). The SMPLX was converted to SMPL format and a skeleton was inferred from the SMPL body shape using the OSSO system (code available open source at github.com/MarilynKeller/OSSO).

The primary advantage of using SMPLX is its ability to accept height/weight BMI regression parameters for desired statures, a function SMPL lacks. The SMPL is needed since it is the only allowed input format of the OSSO system. The body and skeleton models were saved to three-dimensional objects (STL format) and rigidly aligned with the head of the SAFER HBM. The obtained model geometries will hereby be referred to as the *target HBMs*.

### *Pre-processing of the template HBM*

To facilitate the impending morphing procedure, the SAFER HBM was roughly positioned from an occupant, seated position to an upright, standing position using an in-house developed positioning script for PRIMER. The positioning follows the marionette method, partly described by Poulard et al. (Poulard et al. 2015). In short, the model was forced into position in a series of simulations using pulling cables. The cables (formulated as beam elements) were defined with an initial tensile force that forced the movement, and displacement dampers were used to help the model settle during simulation. The cables were applied to prescribe the position and orientation of the head, shoulders, elbows, hands, knees and feet. The technique allows smooth motion of the body parts from the start location to a target location, while maintaining the physiological constraints of the joints. The positioning was done stepwise with a series of simulations to avoid mesh distortion due to large element deformation. At the end of each simulation step, the nodal coordinates of the model were exported for the subsequent step. Manual adjustment of distorted elements and unwanted part intersections and penetrations was needed after positioning. The final positioned HBM had a minimum Jacobian of 0.12, which is comparable of the original occupant SAFER HBM that had a minimum Jacobian of 0.14. Note how the body shape has been distorted, in particular the rear end of the HBM, as seen in Figure 2.

Two separate surface models were created: one model containing the outer surface of the skeleton and one containing the outer skin surface. Both generated surface models were enclosed volumes and were converted to STL file format. The obtained upright SAFER HBM will hereby be referred to as the *template HBM*.

### *Morphing pipeline*

Based on the obtained target and template HBMs, a 50$^{th}$ percentile male and female pedestrian HBM were established following a newly developed image-registration-based mesh morphing framework (Li et al. 2023), outlined in Figure 3. The approach is based on the previous work of Li (2021), who proposed an image-registration-based morphing framework for head models that was later adapted for the application on HBMs (Li et al. 2021, 2023).

The morphing framework incorporates an initial pre-processing to voxelize the HBM shape to binary images. The target and template models were voxelized using an open-source voxelization algorithm (Adam 2023).

Next, a displacement field was obtained using demons registration. The displacement, which was later applied to the template HBM, defines a transformation of the template model to the map onto the target model. Here, a non-linear registration method implemented in 3D Slicer (open-source software available at www.slicer.org) was used (Johnson and Zhao 2022). Deformable or non-rigid image registration algorithms are types of transformations that map one image to another image by finding a displacement field that optimally links or aligns the two images together (Schwarz 2007). The software application used in the present study (Johnson and Zhao 2022) uses diffeomorphic demons algorithms to register a template image onto a target image and provides the resultant deformation fields in written fileformat. The resulting vectors of the deformation field represent the distances between a geometric point in the target image and a point in the template image.

In the current application, the template images constitute the skeleton and outer skin surface of the SAFER HBM that was roughly positioned from seated position. The target images were the 50$^{th}$ percentile skeleton and skin surfaces described in preceding sections of this article. Demons registration was performed to the binary images containing both both skin and skeleton.

The obtained displacement field is defined in the so-called *fixed* image space. Once the deformation field is generated in the above image registration, the total displacement field is applied to the template HBM. With this final step, the HBM assumed the geometry of the desired 50$^{th}$ percentile geometries, without compromising the mesh quality. The morphing was performed excluding the head and feet of the template SAFER HBM. This approach was to maintain the integrity of the KTH head model and to isolate the geometrical correction to the area of interest. The resulting baseline male and female HBMs are illustrated in Figure 1.

### *Evaluating morphing accuracy*

To evaluate the accuracy of the registration, the voxelized image of the template HBM was warped by the inverse of the generated displacement fields. By comparing the warped image with the target image, the morphing accuracy can be quantified.

To measure the spatial overlap between the morphed HBMs and the target geometry, Dice scores (Ou et al. 2014) were calculated. The Dice coefficient can be used to compare the agreement between the morphed images and their corresponding ground truth. Defined as two times the element overlap divided by the total number of elements in both sets, a Dice value of 0 implies no overlap of the PMHS surface and the morphed surface and 1 translates to a perfect overlap:

$$Dice(A, B) = \frac{2|A \cap B|}{|A| + |B|} \qquad (1)$$

Here, $A$ and $B$ are the binary images of the target and morphed image sets, and $|A|$ and $|B|$are their corresponding number of voxels. The number of shared voxels by the two binary image sets is described by $|A \cap B|$.

As an additional accuracy surrogate, the 95$^{th}$ percentile Hausdorff distance (HD95) (Ou et al. 2014) was calculated. HD95 is a metric describing the maximum distance between any point on the template image and its nearest point on the target image, and vice-versa:

$$\mathrm{HD}(M, T) = \max\left(\max_{m \in M} \min_{t \in T} d(m, t), \max_{t \in T} \min_{m \in M} d(m, t)\right) \qquad (2)$$

Here, *d(m, t)* is the Euclidean distance between the spatial locations of two points in the target point set $T$ and morphed point set $M$, while $m$ and $t$ represent a point in the respective sets. To avoid the influence of outliers, the 95$^{th}$ percentile HD was calculated instead of the maximum HD. A small HD95 value indicate a better alignment of the skin and skeleton boundaries between the target and morphed image sets.

## ***Personalized validation of pedestrian HBMs***

To evaluate the whole-body kinematic biofidelity of the SAFER HBM pedestrian, along with the presented personalization pipeline, personalized SAFER HBMs were evaluated by means of experiments using human cadavers conducted by Forman et al. (2015ab). Forman et al. performed full-scale pedestrian impact tests on PMHS struck laterally in a mid-gait stance by a generic vehicle buck. Three tests were documented, involving three different PMHS denoted V2370, V2371 and V2374. The trajectories of the head center of gravity (CoG), pelvis, and T1 and T8 vertebrae were tracked via video and a 3D motion tracking system and kinematic time histories from the head were gathered from accelerometers, and angular rate sensors at selected several anatomical locations. For a comprehensive description of the conducted experiments, the reader is referred to the original publications (Forman et al. 2015ab).

Three SAFER HBMs were processed to match the anthropometries and stances of the PMHS. The presented morphing framework was followed to obtain representative body shapes, sizes and skeletons of the PMHS. The generated baseline male 50$^{th}$ model was used as the template geometry and a SMPL model representative of each PMHS subject's BMI was obtained following the same approach as described in preceding sections of this article. The personalized HBMs were subsequently positioned to a mid-gait stance with the left foot forward and right foot behind the body. Positioning was done using the marionette method. The arms were positioned in front of the body and the hands were coupled using rigid body constraints. The external measurements of the personalized HBMs matched well with the PMHS measurements and a selection of representative dimensions are shown in Figure 1. The Jacobians of the positioned and morphed PMHS were all above 0.12.

The HBMs were struck laterally using a publicly available FE model of a generic sedan buck model developed by Pipkorn et al. (2014). The friction coefficient between the HBM and the buck was set to 0.30. The dimensions of the vehicle were in agreement with the physical buck model used in the experiments, see further details in previous publications (Pipkorn et al. 2014). The impact velocities are presented together with the PMHS and HBM details in Table 1.

The HBMs were modeled with accelerometers at the same anatomical locations as the PMHS. Reaction forces were measured at joints on the buck model, placed at the same locations as the load cells between the load frame and frontal components of the buck in the experiments.

The capability of the SAFER HBM pedestrian to reproduce the trajectories of the anatomical locations, along with the head linear and angular velocities and accelerations was subsequently evaluated, as well as the produced reaction forces of the buck model. The linear acceleration and angular velocity signal was filtered to CFC 1000 and CFC 60 respectively, and all force signals were filtered to CFC 60, in accordance with the experiments.

To objectively evaluate the correlation of the simulation kinematic time-histories of the head COG with the PMHS head accelerometer, the CORrelation Analysis (CORA) rating method was used. The CORA score indicates how well two time-histories correlate and ranges between 0 and 1, where 1 indicates a perfect correlation and 0 a poor correlation. In general, a CORA score above 0.44 indicates a "fair" biofidelity, and above 0.68 is considered as "good" biofidelity. The default parameters presented in the CORA manual were used (Thunert 2017). The presented scores are calculated as the average of each X-, Y-, and Z-component score.

## Results

### *Generated HBMs*

In Table 2, the Dice scores and HD95 distances are presented for baseline HBMs and the three personalized HBMs. The average Dice score was 0.93 and average HD95 distance was 16.0 mm. To provide an indication of the element quality after morphing, the element minimum Jacobian for each generated model is included. The minimum Jacobian was greater than 0.10 for all HBMs.

For further illustration of the morphing accuracy, the model-to-model distance of the generated baseline models are presented in Figure 4. The model-to-model distance is the computed absolute Hausdorff distance between the target models and morphed models at each point. Any part intersections or negative Jacobians that were introduced after the morphing procedure were post-processed and repaired if required.

### *Trajectories and Kinematics*

The trajectories of the simulated impact and the corresponding experimental impacts are presented in Figure 5. The two-dimensional trajectories are presented relative to the buck model. The trajectories of the head, spine and pelvis were close to the trajectories documented in the corresponding PMHS experiments. One key difference between the trajectories of the PMHS and the personalized HBMs was found in the pelvis impact sequence. For the HBMs in the simulation, the pelvis went over the hood and lost contact with it, but for the PMHS models, the contact between the pelvis and hood was continuous. This is seen in Figure 6, where the impact sequence is shown in a posterior view of the HBM and PMHS in a series of time steps. The pelvis movement is also highlighted by the corresponding trajectory (see Figure 5, Pelvis around X = 500 mm). Snapshots of pedestrian impact sequences for subjects V2370 and V2371 are available in the Appendix.

In all three impact tests, the subject-specific HBMs showed a delay in the impact sequence, with the final head impact occurring later compared to the experiments. The head contact occurred between 3 and 10 ms later compared to the experiments. The wrap-around-distance (WAD) was not less than 23 mm away from the experimentally derived WAD in all three impacts. The simulated head contact time and WAD for each impact test are presented in Table 3.

The head CoG angular velocity and linear acceleration time-histories are presented jointly with corresponding CORA ratings in Figure 7. The average CORA rating for the angular velocity was 0.70, and 0.63 for the linear acceleration. Resultant forces from the lower and upper bumper, grille and hood of the buck model showed good correlation with the PMHS experiments. Force time-histories are included in the Appendix.

The experimental results are included in all mentioned figures. The same coordinate system of the head and buck model was used as in the experimental study (Forman et al. 2015ab). Trajectories and time-histories are all truncated from the time of initial leg impact.

## Discussion

In this study, a 50th percentile male and female pedestrian SAFER HBM have been established using state-of-the-art image-registration-based morphing techniques and positioning tools. In conjunction with the baseline HBMs, an HBM personalization framework has been introduced. The presented framework allows for rapid and landmark-free generation of subject-specific HBMs, and accounts for the body size and shape as well as the skeletal morphology. The baseline HBMs together with the personalization framework was evaluated in simulated PMHS impact test, where the biofidelity of the models was demonstrated. The baseline HBMs and personalization framework can be used to study head injury mechanisms in reconstructed pedestrian-to-vehicle collisions, among many other applications.

### *Validity of Baseline HBMs*

Two baseline 50th percentile HBMs have been established by morphing a template HBM to a target geometry, without accounting for any other biological differences apart from the human anthropometry. Yet, human attributes have been argued to have significant effects on the material properties (Hwang et al. 2016). Such variations among populations were neglected in this study, yet judging by the CORA ratings, the HBM seem to exhibit satisfactory biofidelity in the lateral impacts investigated in this study. This is in line with the findings of Hwang et al. (2016), who found that the geometry and posture of the HBM have larger effects than the material properties of the human bone and soft tissues looking at occupant impact responses.

The two baseline models are based on a seated, occupant HBM that was positioned into an upright stance using FE simulation. The subsequent morphing procedure only takes the body shape and skeletal morphology into account. During the process of establishing the baseline pedestrian HBMs, deformations of the internal parts of the HBM were introduced conjointly with the positioning and applied displacement fields. Previous studies have found significant differences in abdominal organ position and shape between supine and seated postures (Beillas et al. 2009). With this in mind, the positioning from a seated to an upright stance might introduce unintended errors from the displaced positions, morphed volumes and changed mass distributions of the inner organs. A qualitative check was performed after morphing and the inner organs remained intact in mesh quality and the shape was not heavily distorted compared to the original occupant HBM. Neither new intersections in the contact definitions were detected. And most importantly, the validation presented in this study show that the biofidelity was preserved after the positioning and morphing procedures. Even though further validation might be needed, the results of this study indicates a good preservation of the biofidelity of the HBM, despite any possible mesh changes of the internal organs.

As indicated by the Dice and HD95 values, the female baseline HBM was not as close to its target shape as the male baseline HBM. The Dice value was lower and HD95 was higher for the female baseline model compared to the male HBM. This is explained by the obvious similarities between the $50^{th}$ percentile male template and the target $50^{th}$ percentile male. Very little displacement of the mesh was needed to morph the male template HBM to match the target geometry. The female target image on the other hand, differs significantly in both size and shape compared to the template $50^{th}$ percentile male. Thus, more distortion of the mesh is needed, resulting in less morphing accuracy. As seen by Figure 4, the absolute error of the female baseline model is concentrated at the soft tissues that differ the most from the male HBM, i.e. chest region, lower abdomen and rear end.

### *Personalization framework*

The SAFER HBM sustained significant deformation during positioning due to the leg rotations around the pelvis when transforming a sitting occupant model to a standing upright posture. This resulted in some mesh distortions that were manually repaired. Any part intersections or negative Jacobians introduced by the positioning were manually post-processed and mesh quality was restored. The morphing procedure introduced no new intersections or negative Jacobians, which stresses the robustness of the proposed image-registration-based personalization tool. Once the baseline male and female HBMs were established, subject-specific HBMs were generated with the baseline HBMs as template models. No post-processing was necessary after the personalization procedure. This means that generating subject-specific HBMs based on the established baseline HBMs require minimal effort and little pre-and post-processing time.

The HBM personalization procedure described in this study differs from previously proposed personalization techniques. Many studies prior to this have introduced non-rigid deformation techniques to adapt the geometry of HBMs (Larsson et al. 2022; Poulard et al. 2016; Chen et al. 2018; Hwang et al. 2016). The presented personalization-by-morphing approaches have often been limited to isolated body parts, e.g. the femur (Klein et al. 2015), pelvis (Salo et al. 2015) or the head (Li 2021). Proposed morphing techniques of HBMs seldom take into account both skeletal and outer shape morphology simultaneously. For instance, Chen et al. (2018) used a morphing method using THUMS HBM, which did not account for the external surface of the HBMs. They developed subject-specific

HBMs by morphing an existing HBM to CT and anthropometric measurements, following a so-called dual kriging interpolation process (Poulard et al. 2016) using 144 control points constructed from Computer Tomography (CT) scans and anthropometric measurements. In contrast to the personalization framework presented in this study, this kriging interpolation process is very reliant on geometry data (CT or anthropometric measurements of subject). The current proposed morphing framework requires minimal information about the subject anthropometry apart from the subject height and weight. Any additional information, such as subject age or additional CT scans, can however be used to increase the HBM personalization accuracy even further.

In this study, a method of mesh morphing has been proposed that introduces very few practical limitations. As mentioned above, the proposed personalization framework does not consider the inner organs and soft tissues (although they are also affected by the morphing). Preparing other kinds of template models, it is though theoretically possible to adapt the morphing process to also morph the subject geometry to specific inner organs. For instance, the method works well with generating subject-specific anatomically detailed head models (Li 2021). The methodology presented in this study is neither exclusive to the SAFER HBM. The same image-registration-based morphing procedure can theoretically be used to morph any other existing FE HBM. Some of the many applications of the presented morphing is investigated in previous publications (Li et al. 2023).

### *Validation*

This study focused on the development of baseline SAFER HBM pedestrians and the development of a personalization framework. To show the robustness of the morphed models in crash simulations, we reconstructed three PMHS experimental impact tests. To demonstrate the personalization framework, the trajectories of the HBMs were evaluated by means of individual responses instead of the proposed normalized PMHS responses. Three specific experiment scenarios were reconstructed, and the simulated trajectories of the HBMs exhibited very good agreement with PMHS, with nearly identical upper body trajectories for each location in Y-Z plane. The CORA scores of the kinematic time-histories indicate a fair agreement between experiment and simulation time-histories (Thunert 2017), demonstrating that the HBM was capable of mimicking the response of a human body in lateral impacts. Of course, there is room for improvement.

First of all, the accuracy of the reconstructed PMHS experiments could be further improved by using more representative template body models. In this study, a morphing template image was chosen based on the BMI, age and sex of each PMHS. The generated template image is thus a generic representative of a human body with the wanted BMI, age and sex. Of course, this will not be an exact representation of the PMHS, due to the many possible subject-to-subject variations. Knowing the PMHS measurements, even better target geometry can be generated. With a perfect target model, the distance between the head, T1 vertebrae, T8 vertebrae and pelvis could be much more accurate. As seen in the simulated trajectories, these landmarks, especially the pelvis, do not perfectly coincide with the PMHS. Achieving better accuracy is entirely feasible using the presented image-registration morphing framework.

Secondly, the accuracy of the reconstructed PMHS experiments can be further improved by more accurate positioning of the HBMs. Under the assumption that the arms have little effect on the impact kinematics, the arms of the three subject-specific PMHS were uniformly placed in front of the body. The arms and shoulders could have been positioned with better accuracy, as well as capturing the shrugged posture seen in the PMHS due to the support of the cadavers in the experiments better. Furthermore, one PMHS (V2374) was missing an arm, which was not accounted for by the subject specific HBM. Improving the HBM-to-PMHS similarity would most likely increase the agreement between the simulations and experiments. It should nonetheless be mentioned, that positioning of the HBMs using the simulation-based marionette method is a time consuming process, especially regarding fine tuning of landmark positions.

Lastly, there might also be inherent errors from the used FE buck model. The buck FE model has previously been validated in regard to car-pedestrian interactions (Takahashi et al. 2014). In general, the buck FE model has shown good correspondence with the mechanical buck. But some discrepancies have been found in the hood edge reaction force (Takahashi et al. 2014). Nevertheless, the extracted forces from the buck joints exhibited good agreement with PMHS experiments. Evaluating the validity of the buck FE model is however beyond the scope of this study.

The simulated trajectory of the pelvis showed less similarity to the PMHS experiment. This might be explained by the extra weight of the experimental equipment, which were accounted for in the simulations by distributing the extra weight (7.4 kg) equally over the HBM body parts. In the PMHS experiments, cables and equipment were put in a bag attached to the lower back of the PMHS. The mass distribution of the equipment will influence the kinematics of the PMHS and therefore, in the HBM validations, concentrating the mass of the cables to the lower back/buttocks can improve the agreement between model predictions and test results, especially the pelvis.

The impact response of the HBMs were delayed compared to the PMHS experiments. Wu et al. (2017), who performed the same validation simulations using scaled THUMS HBMs, observed the same phenomena. The SAFER HBM is originally based on the THUMS v3, which could be a reason to the similar impact response discrepancy. Wu et al. (2017) hypothesized that the delay could be due to a softer neck in terms of lateral bending stiffness of the THUMS model compared to the PMHS. This could be the case for the SAFER HBM as well since the response of the thoracic spine appears to be more similar in respect of timing to the PMHS experiments for subjects V2371 and V2374.

### *The importance of personalization*

In this study, we did not explicitly evaluate whether subject-specific morphing increases the biofidelity of pedestrian impact kinematics. Either way, it is a very important question: is geometrical personalization really necessary? In 2016, Poulard et al. (2016) attempted to answer this question by performing PMHS simulations using one generic HBM, one globally scaled HBM and one personalized (morphed) HBM. In the study, the generic HBM was represented by a 50$^{th}$ percentile male THUMS HBM. The globally scaled HBMs were obtained by applying a global scaling factor to match the height of the PMHS and a second factor to scale the HBM in the transverse plane to match the PMHS weight. The morphed HBM was obtained using the Dual Kriging interpolation process (Trochu 1993). The simulations were set up to correspond to PMHS experiments, where two male obese PMHS were struck laterally by a mid-sized sedan travelling at 40 km/h, which is very similar to the validation experiments used in the present study. When comparing the simulations against the cadaver experiments, Poulard et al. (2016) could clearly demonstrate how the subject-specific HBMs obtained by morphing produced the best correlation to the cadaver experiments. The morphed HBMs resulted in better prediction of the WAD, impact locations and timing. The CORA scores for the generic, globally scaled and morphed models were 0.85, 0.92 and 0.95 respectively, serving as a clear indicator of an increase in biofidelity by using subject-specific HBMs. Similar conclusions were drawn in previous studies by Piqueras-Lorente et al. (2018), Larsson et al. (2019, 2022), Hwang et al. (2016) using occupant HBMs.

## Conclusions and future work

In this study, a 50$^{th}$ percentile male and female pedestrian SAFER HBM have been established using state-of-the-art image-registration-based morphing techniques and positioning tools. In conjunction with the baseline HBMs, an HBM personalization framework has been introduced. The presented framework allows for rapid and landmark-free generation of subject-specific HBMs, and accounts for the body size and shape as well as the skeletal morphology. A fair agreement was demonstrated regarding kinematics under qualitative assessment in simulated PMHS impacts. The trajectories of the upper body were nearly identical to those seen in PMHS experiments. The established baseline female and male HBMs and personalization framework can be used to study head injury mechanisms in reconstructed pedestrian-to-vehicle collisions, among many other applications.

## Conflict of interest statement

The authors declare no conflict of interest.

# Acknowledgements

This study was partly financed by Swedish Governmental Agency for Innovation Systems (Vinnova) (no. 2019-03386) and the Swedish Research Council (VR) (no. 2020-04724 and 2020-04496). The computations were enabled by resources provided by the Swedish National Infrastructure for Computing (SNIC), partially funded by the Swedish Research Council through grant agreement no. 2018- 05973.

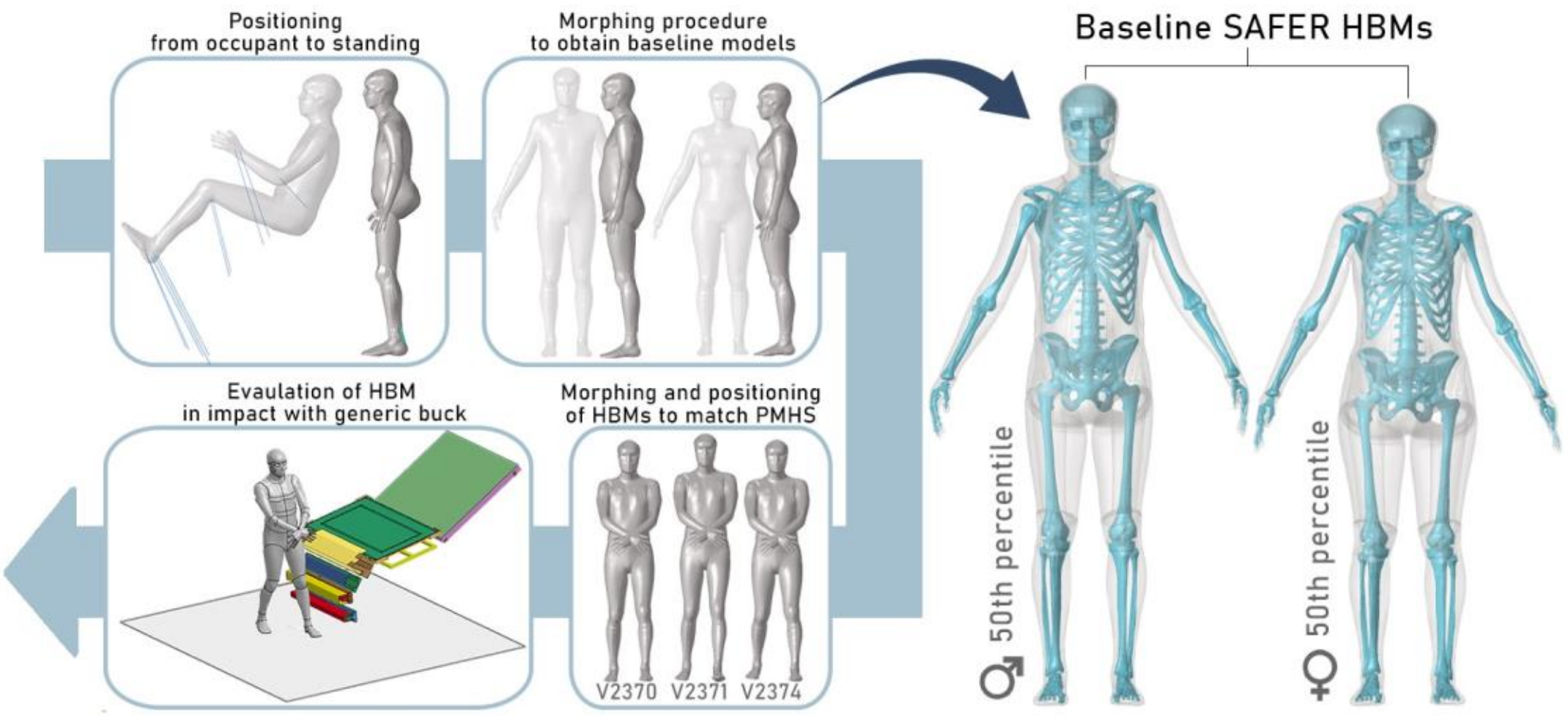

Figure 1. Study overview and presentation of baseline SAFER HBMs: The occupant HBM was computationally positioned to an upright standing position by using pulling cables. The upright HBM was subsequently morphed to obtain the body shape and skeleton representative of a 50$^{th}$ percentile male and female, constituting as the newly established baseline SAFER HBM pedestrians. To evaluate the pedestrian HBM and personalization technique, three HBMs were prepared with personalized anthropometries and were subjected to lateral impacts using a generic buck model. The results were evaluated my means of post mortem human subjects (PMHS) experimental data. *[Two-column figure]*

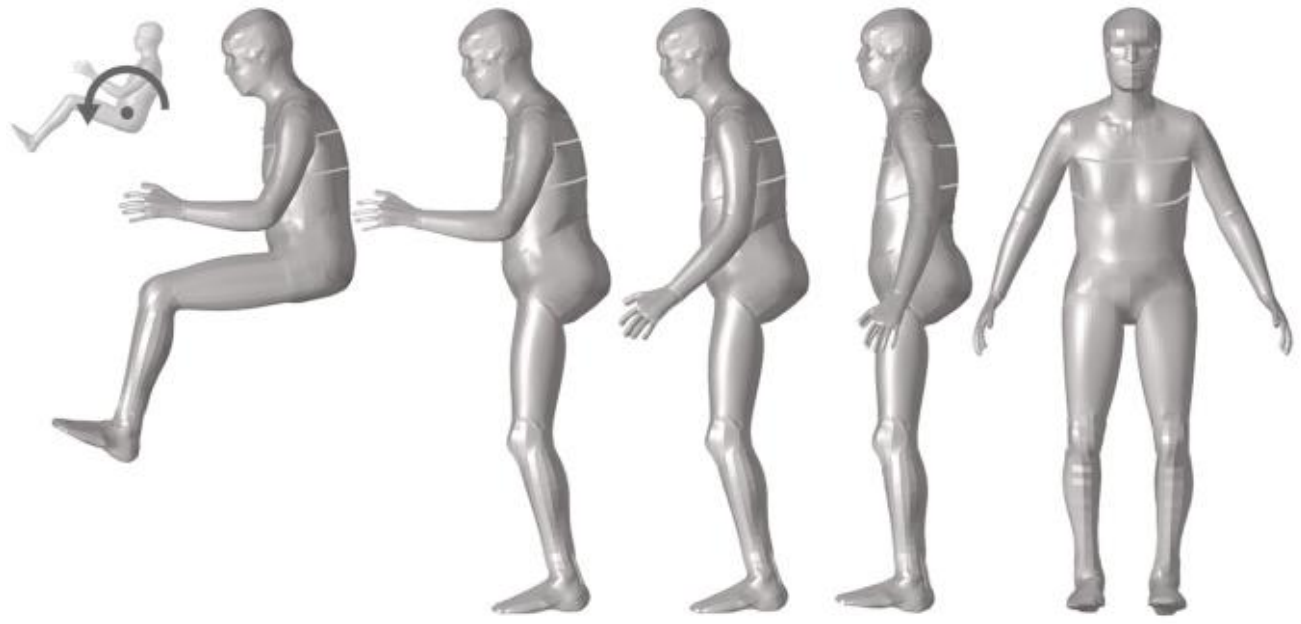

Figure 2. Illustrations of a selection of the positioning steps to obtain an upright SAFER HBM. *[One-column figure]*

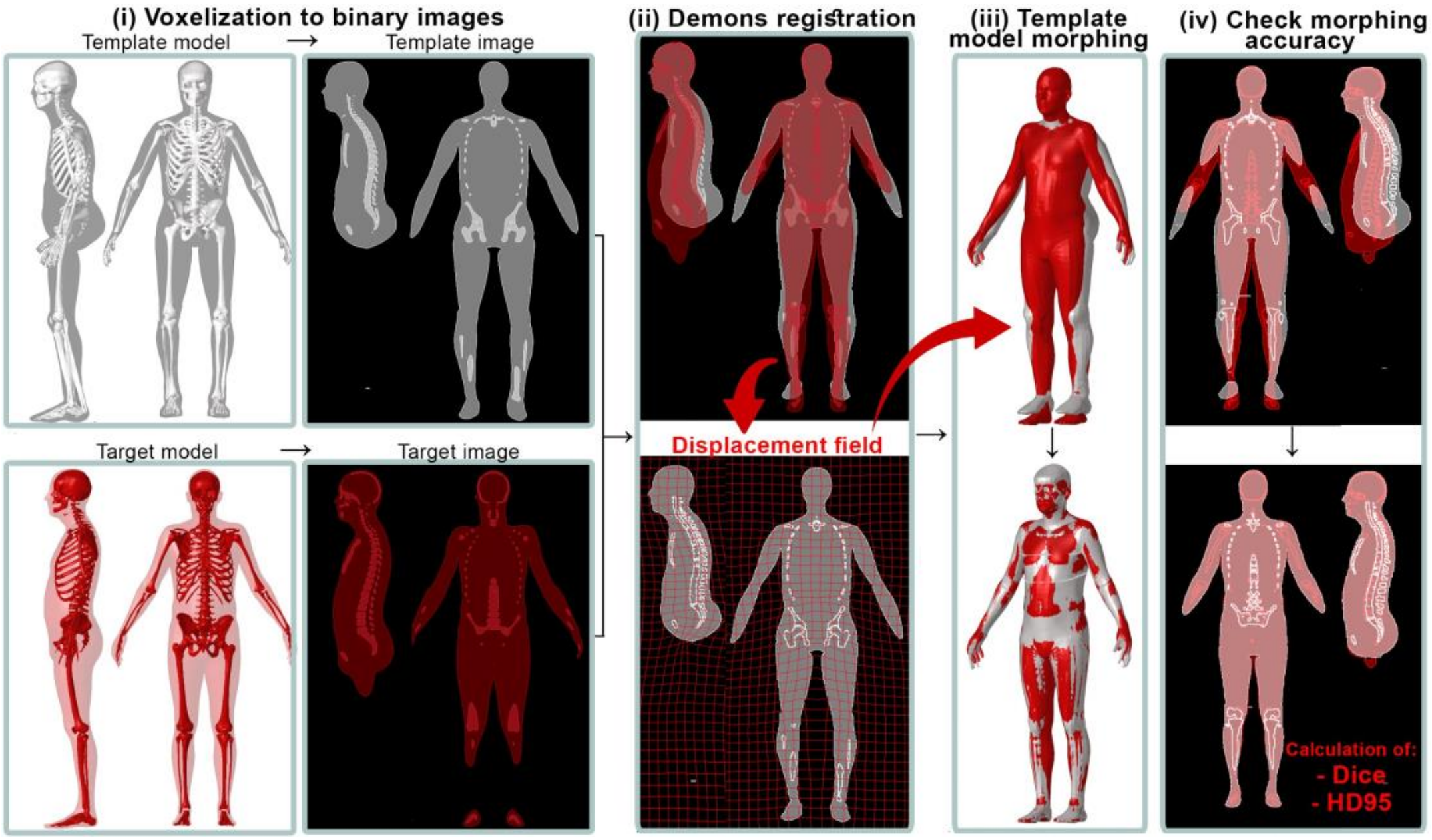

Figure 3. Basic morphing framework using a target and template HBM. (i) The target and template outer skin and skeleton are voxelized to binary images and rigidly aligned by the head. (ii) Demons registration using the template binary image set as the target images and the target binary image set as moving to generate a displacement field defined on the fixed image space. (iii) The displacement field is applied to the template HBM, morphing the HBM to the desired personalized geometry. (iv) The image set of the template HBM is hardened by the inverse of the generated displacement field and is subsequently compared with the target image set to quantify personalization accuracy using Dice and HD95 metrics. *[Two-column figure]*

Table 1. Anthropometric details and measurements of the positioned HBMs. All measurements are in mm if not stated otherwise. The PMHS measurements from the experimental study are provided within parenthesis and all height measurements are measured from the buck ground reference plane. Impact velocities are added for reference.

*7.4 kg was added to account for the added equipment used in PMHS tests. The extra weight was distributed evenly over the HBM flesh. *[Two-column figure]*

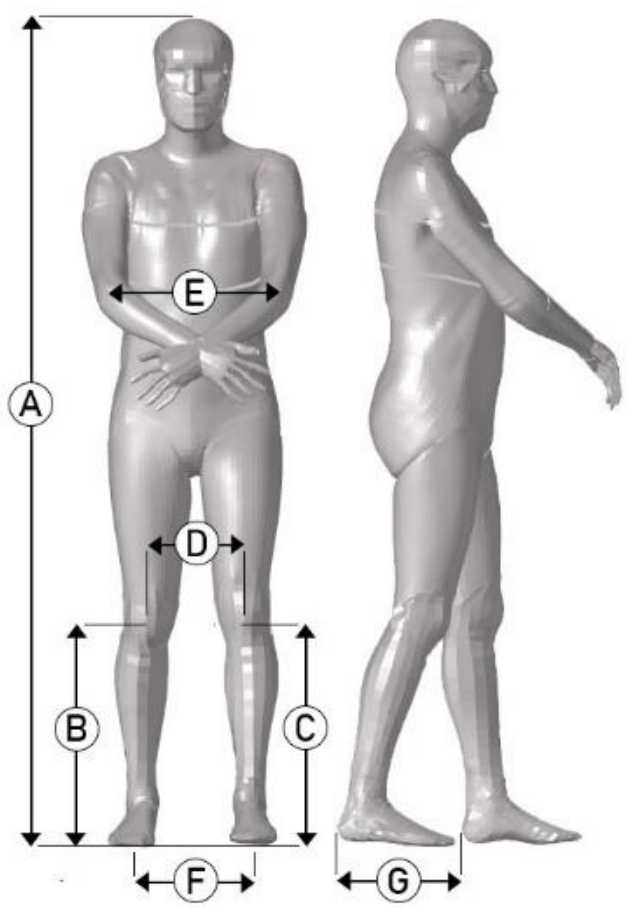

| | | V2370 | V2371 | V2374 |
|---|---|---|---|---|
| | Impact velocity | 40.0 km/h | 39.4 km/h | 40.2 km/h |
| | Age | 73 years | 54 years | 67 years |
| | Weight* | 72.6 kg | 81.6 kg | 78.0 kg |
| | Height (supine) | 1795 | 1870 | 1780 |
| A | Height (positioned) | 1785 *(1779)* | 1823 *(1846)* | 1793 *(1778)* |
| B | Knee Height Left | 494 *(494)* | 492 *(492)* | 503 *(503)* |
| C | Knee Height Right | 504 *(494)* | 501 *(514)* | 519 *(523)* |
| D | Knee to Knee | 209 *(202)* | 208 *(219)* | 231 *(228)* |
| E | Elbow to Elbow | 380 *(382)* | 380 *(562)* | 378 *(412)* |
| F | Heel to Heel Front | 252 *(255)* | 245 *(239)* | 298 *(291)* |
| G | Heel to Heel Side | 267 *(270)* | 381 *(382)* | 379 *(381)* |

Table 2. Minimum Jacobians, Dice values and HD95 distances for the baseline HBMs and personalized HBMs (PMHS). *[One-column figure]*

| | Min. Jacobian [-] | Dice [-] | HD95 [mm] |
|---|---|---|---|
| Baseline Male | 0.14 | 0.95 | 12.7 |
| Baseline Female | 0.12 | 0.82 | 19.6 |
| V2370 | 0.12 | 0.97 | 21.1 |
| V2371 | 0.11 | 0.97 | 13.7 |
| V2374 | 0.10 | 0.95 | 12.7 |

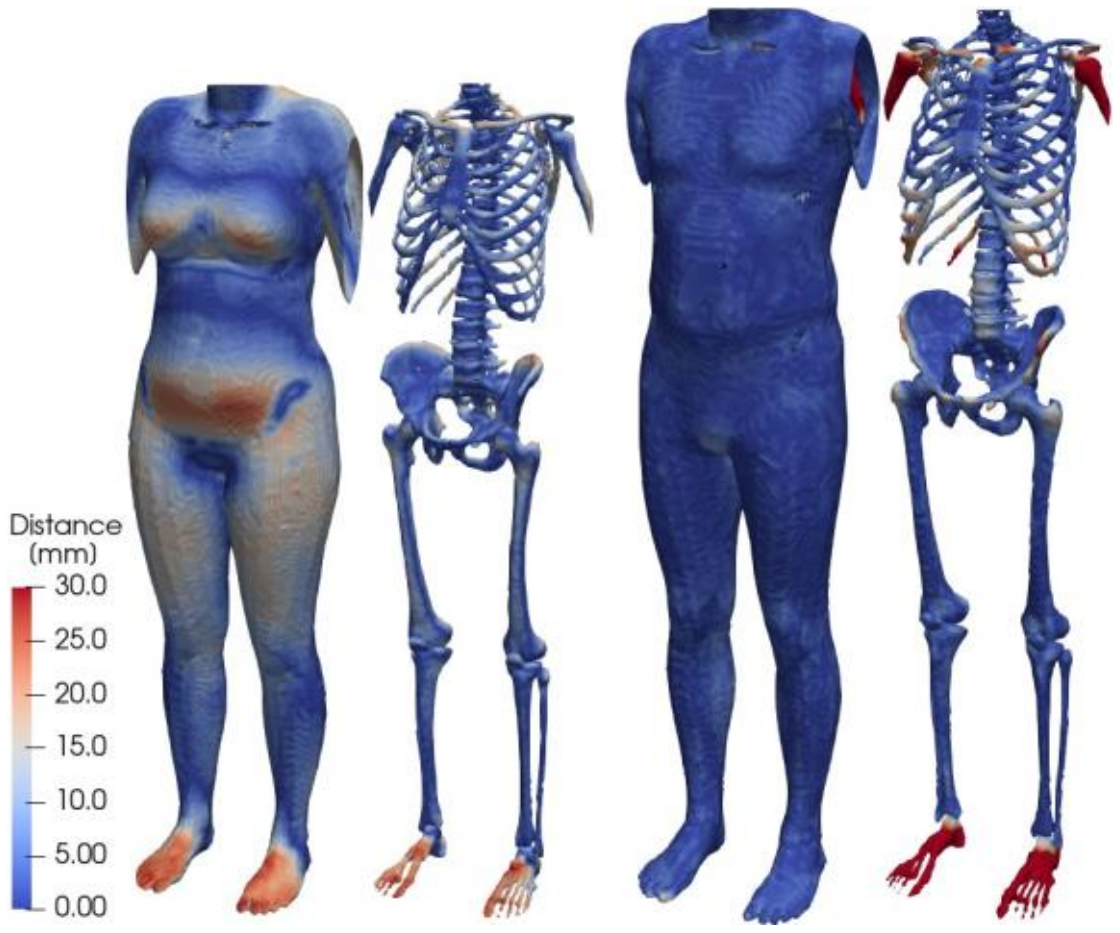

Figure 4. Model-to-model absolute distance of female baseline outer skin and skeleton (left) and male baseline outer skin and skeleton (right). Red indicates a larger absolute error, while blue shows a perfect alignment of template-to-target points. *[One-column figure]*

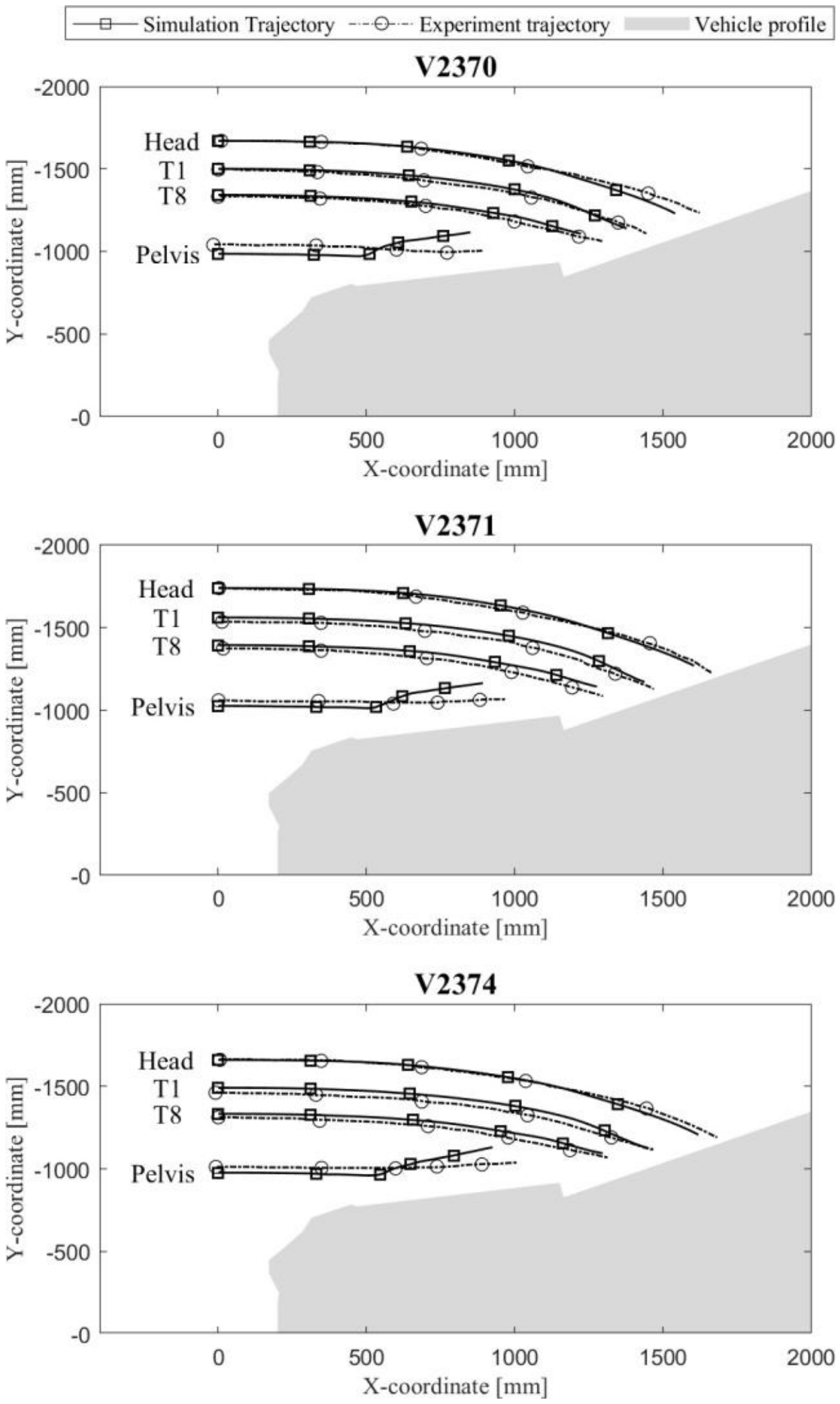

Figure 5. Trajectories of the personalized HBMs in lateral impacts. Markers are displayed for every 30 ms from the point of leg contact (t = 0). *[One-column figure]*

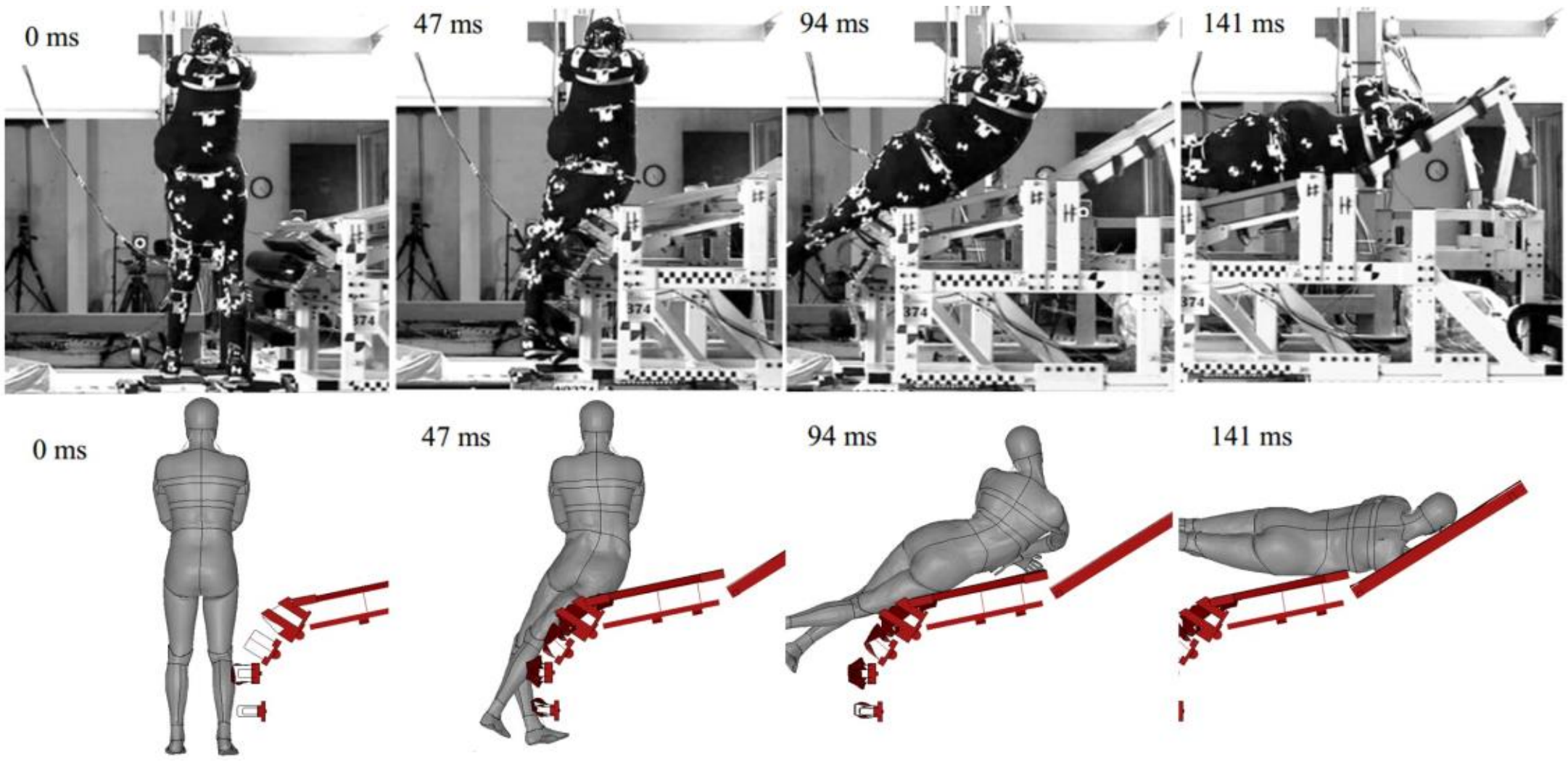

Figure 6. Posterior view of V2374 in selected time steps. Images of PMHS experiments are extracted from a previous publication by Wu et al. (2017). *[Two-column figure]*

Table 3. Head contact times and WADs. The presented impact times are relative to the time of leg contact. The WAD was measured from the buck ground reference plane. Experimental measurements are presented in parentheses. *[One-column figure]*

| | Head contact time [ms] | Head WAD [mm] |
|---|---|---|
| V2370 | 142 *(135)* | 2167 *(2190)* |
| V2371 | 148 *(138)* | 2171 *(2175)* |
| V2374 | 144 *(141)* | 2190 *(2200)* |

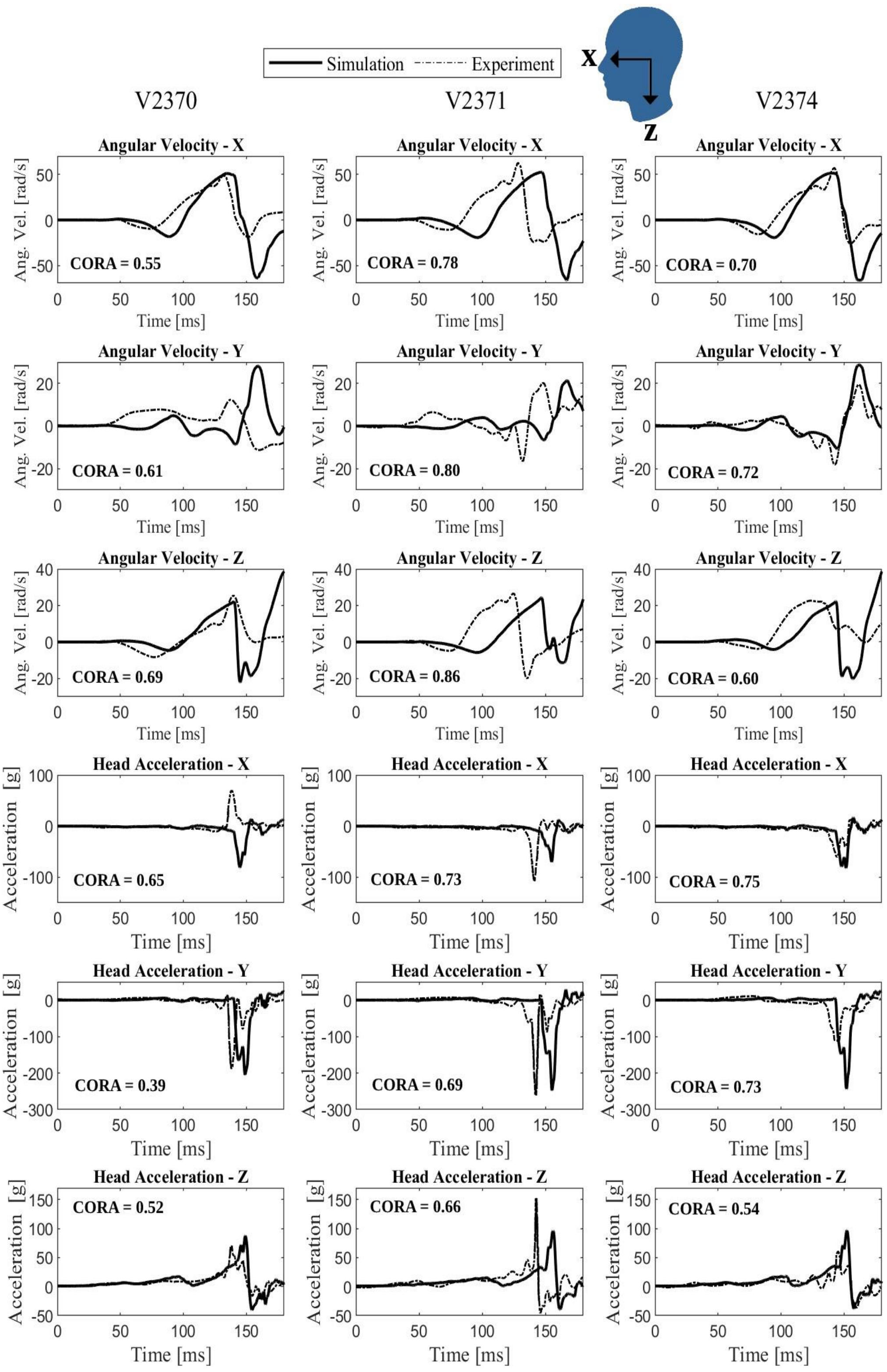

Figure 7. Head CoG acceleration and angular velocity time-histories.